\documentclass[10pt,twocolumn,letterpaper]{article}

\usepackage{wacv}
\usepackage{times}
\usepackage{epsfig}
\usepackage{graphicx}
\usepackage{amsmath}
\usepackage{amssymb}
\usepackage{float}
\usepackage{subfigure}
\usepackage{array}
\usepackage{multirow}
\usepackage{soul}
\usepackage{authblk}

\wacvfinalcopy % *** Uncomment this line for the final submission

\usepackage[pagebackref=true,breaklinks=true,letterpaper=true,colorlinks,bookmarks=false]{hyperref}

\newcommand{\PreserveBackslash}[1]{\let\temp=\\#1\let\\=\temp}
\newcolumntype{C}[1]{>{\PreserveBackslash\centering}p{#1}}

\def\comm[#1]{{\small \textcolor{red}{\emph{}}}}
\def\red[#1]{{\textbf{#1}}}
\def\redn[#1]{{#1}}
\def\blue[#1]{{#1}}
\def\green[#1]{\textcolor{green}{#1}}
\def\revise[#1]{{\small \textcolor{blue}{\emph{#1}}}}
\def\del[#1]{\setstcolor{red} \st{}}

\newcommand\ken[1]{{\small \textcolor{red}{\emph{}}}}

\def\wacvPaperID{30} % *** Enter the wacv Paper ID here

\makeatletter
\renewcommand\AB@affilsepx{, \protect\Affilfont}
\makeatother

\ifwacvfinal\fi
\begin{document}

%%%%%%%%% TITLE
\title{Face Sketch Synthesis with Style Transfer using Pyramid Column Feature}

% Authors at the same institution
\author{Chaofeng Chen$^1$\thanks{indicates equal contribution}~, Xiao Tan$^{2*}$\thanks{This work was done when Xiao Tan was a postdoc at HKU.}~~, and Kwan-Yee K. Wong$^1$ \\
$^1$The University of Hong Kong, $^2$Baidu Research \\
{\tt\small \{cfchen, kykwong\}@cs.hku.hk, tanxchong@gmail.com}
}

\maketitle
\ifwacvfinal\thispagestyle{empty}\fi

%%%%%%%%% ABSTRACT
\begin{abstract}

In this paper, we propose a novel framework based on deep neural networks for face sketch synthesis from a photo. Imitating the process of how artists draw sketches, our framework synthesizes face sketches in a cascaded manner. A content image is first generated that outlines the shape of the face and the key facial features. Textures and shadings are then added to enrich the details of the sketch. We utilize a fully convolutional neural network (FCNN) to create the content image, and propose a style transfer approach to introduce textures and shadings based on a newly proposed pyramid column feature.
We demonstrate that our style transfer approach based on the pyramid column feature can not only preserve more sketch details than the common style transfer method, but also surpasses traditional patch based methods. Quantitative and qualitative evaluations suggest that our framework outperforms other state-of-the-arts methods, and can also generalize well to different test images. Codes are available at \url{https://github.com/chaofengc/Face-Sketch}

\end{abstract}

%%%%%%%%% BODY TEXT
%==========================================================================
\section{Introduction}
Face sketch synthesis has drawn great attention from the community in recent years because of its wide range of applications. For instance, it can be exploited in law enforcement for identifying suspects from a mug shot database consisting of both photos and sketches. Besides, face sketches have also been widely used for entertainment purpose. For example, filmmakers could employ face sketch synthesis technique to ease the cartoon production process.

\begin{figure}[t]
\centering
\subfigure[Photo]{
\label{fig:example_a}\includegraphics[width=0.23\linewidth]{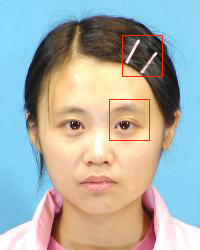}}
\subfigure[MRF \cite{wang2009face}]{
\label{fig:example_b}\includegraphics[width=0.23\linewidth]{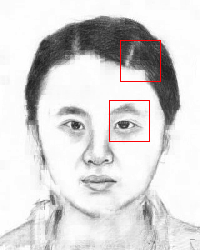}}
\subfigure[WMRF \cite{zhou2012markov}]{
\label{fig:example_c}\includegraphics[width=0.23\linewidth]{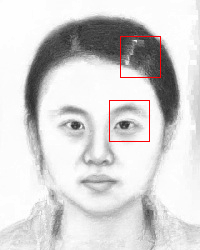}}
\subfigure[SSD \cite{song2014real}]{
\label{fig:example_d}\includegraphics[width=0.23\linewidth]{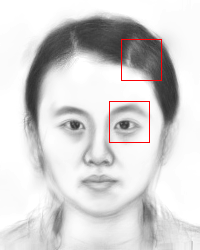}}
\subfigure[FCNN \cite{zhang2015end}]{
\label{fig:example_e}\includegraphics[width=0.23\linewidth]{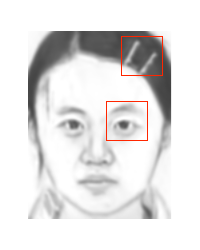}}
\subfigure[BFCN \cite{zhang2017content}]{
\label{fig:example_f}\includegraphics[width=0.23\linewidth]{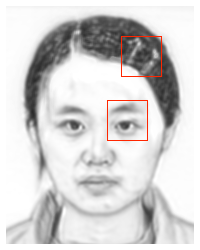}}
\subfigure[\cite{gatys2015neural}$^*$]{
\label{fig:example_g}\includegraphics[width=0.23\linewidth]{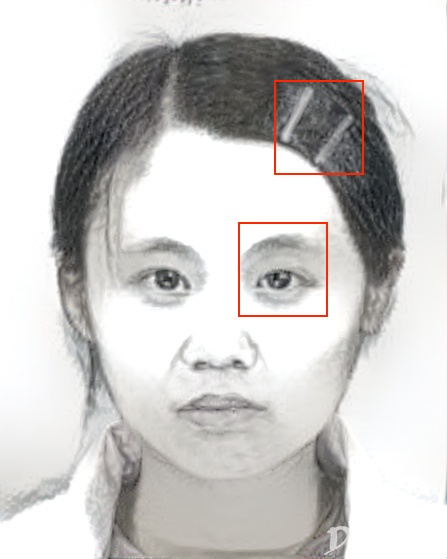}}
\subfigure[Ours]{
\label{fig:example_h}\includegraphics[width=0.23\linewidth]{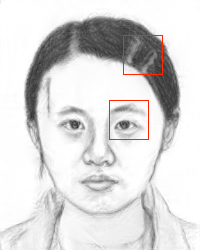}}

\begin{minipage}[t]{1\linewidth}
\centering
\includegraphics[width=0.11\linewidth]{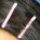}
\includegraphics[width=0.11\linewidth]{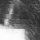}
\includegraphics[width=0.11\linewidth]{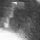}
\includegraphics[width=0.11\linewidth]{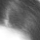}
\includegraphics[width=0.11\linewidth]{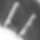}
\includegraphics[width=0.11\linewidth]{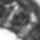}
\includegraphics[width=0.11\linewidth]{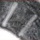}
\includegraphics[width=0.11\linewidth]{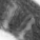}
\end{minipage}
\begin{minipage}[t]{1\linewidth}
\centering
\includegraphics[width=0.11\linewidth]{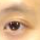}
\includegraphics[width=0.11\linewidth]{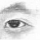}
\includegraphics[width=0.11\linewidth]{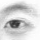}
\includegraphics[width=0.11\linewidth]{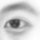}
\includegraphics[width=0.11\linewidth]{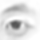}
\includegraphics[width=0.11\linewidth]{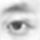}
\includegraphics[width=0.11\linewidth]{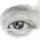}
\includegraphics[width=0.11\linewidth]{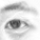}
\end{minipage}

\begin{minipage}[t]{1\linewidth}
\footnotesize
\centering
\begin{tabular}{*8{C{0.6cm}}}
(a) & (b) & (c) & (d) & (e) & (f) & (g) & (h)
\end{tabular}
\end{minipage}

\caption[Caption for LOF]{Face sketches generated by existing methods and the proposed method. Our method can not only preserve both hair and facial content, but also contains sharp textures. %
(Note that (g) is obtained from the deep art website\setcounter{footnote}{0}\footnotemark~using the photo as content and a sketch from the training set as style. For results of different sizes, we simply padded them with zeros.)}
\label{fig:example_comp}
\end{figure}
\footnotetext{\url{https://deepart.io/}} 

Unfortunately, there exists no easy solution to face sketch synthesis due to the big stylistic gap between photos and sketches. In the past two decades, a number of exemplar based methods~\cite{wang2009face,song2014real, zhang2010lighting,zhou2012markov} were proposed. In these methods, a test photo is first divided into patches. For each test patch, a candidate sketch patch is identified by finding the most similar photo patch in a training set of photo-sketch pairs. The main drawback of this approach is that if there exists no photo patch in the training set which is  sufficiently similar to a test patch, loss of content will be observed in the synthesized sketch. For example, the sketches in the first row of Fig.~\ref{fig:example_comp} fail to keep the hairpins. Besides, some methods \cite{song2014real,zhou2012markov} blur away the textures when they try to eliminate the inconsistency between neighboring patches. Another common problem is that the synthesized sketch may not look like the test photo (see the left eye in Fig.~\ref{fig:example_b}). Recently, approaches \cite{zhang2017content,zhang2015end} based on convolutional neural network (CNN) were developed to solve these problems. Since they directly generate sketches from photos, structures and contents of the photos can be maintained. However, the pixel-wise loss functions adopted by these methods will lead to blurry artifacts (see Fig.~\ref{fig:example_e} and ~\ref{fig:example_f}) because they are incapable of preserving texture structures. The popular neural style transfer provides a better solution for texture synthesis. However, there exist two obstacles in directly applying such a technique. First, the brightness of the result is easily influenced by the content of the style image (see the face in Fig.~\ref{fig:example_g}). Second, it requires a style image to provide the global statistics of the textures. If the given style image does not match with the target sketch (which we do not have), some side effects will occur (see the nose in Fig.~\ref{fig:example_g}).

For an artist, the process of sketching a face usually starts with outlining the shape of the face and the key facial features like the nose, eyes, mouth and hair. Textures and shadings are then added to regions such as hair, lips, and bridge of the nose to give the sketch a specific style. Based on the above observation, and inspired by neural style transfer \cite{gatys2015neural}, we propose a new framework for face sketch synthesis from a  photo that overcomes the aforementioned limitations. In our method, a content image that outlines the face is generated by a feed-forward neural network, and textures and shadings are then added using a style transfer approach. Specifically, we design a new architecture of fully convolutional neural network (FCNN) composed of inception layers \cite{szegedy2015going} and convolution layers with batch normalization~\cite{Sergey2015batch} to generate the content image (see Section~\ref{sec:content_net}). To synthesize the textures, we first divide the target sketch into a grid. For each grid cell, we compute a newly proposed pyramid column feature using the training set (see Section~\ref{subsec:pyramid_feature_column}). A target style can then be computed from a grid of these pyramid column features, and applied to the content image. Our approach is superior to the current state-of-the-art methods in that
\begin{itemize}
\item It is capable of generating more stylistic sketches without introducing over smoothing artifacts.
\item \redn[It preserves the content of the test photo better than current state-of-the-art methods.]
\item \redn[It achieves better results in the face sketch recognition task.]
\end{itemize}

%==========================================================================
\section{Related Work}\label{sec:related_work}

\subsection{Face Sketch Synthesis}
Based on the taxonomy of previous studies~\cite{song2014real,zhou2012markov}, face sketch synthesis methods can be roughly categorized into profile sketch synthesis methods~\cite{berger2013style,chen2001example,xu2008hierarchical} and shading sketch synthesis methods~\cite{liu2005nonlinear,song2014real,tang2003face,wang2009face,zhang2015end,zhang2010lighting,zhou2012markov}. Compared with profile sketches, shading sketches are more expressive and thus more preferable in practice. Based on the assumption that there exists a linear transformation between a face photo and a face sketch, the method in~\cite{tang2003face} computes a global eigen-transformation for synthesizing a face sketch from a photo. This assumption, however, does not always hold since the modality of face photos and that of face sketches are quite different. Liu \etal~\cite{liu2005nonlinear} pointed out that the linear transformation holds better locally, and therefore they proposed a patch based method to perform sketch synthesis. In \cite{wang2009face}, a MRF based method was proposed to preserve large scale structures across sketch patches. Variants of the MRF based methods were introduced in~\cite{zhang2010lighting,zhou2012markov} to improve the robustness to lighting and pose, and to render the ability of generating new sketch patches. In addition to these MRF based methods, approaches based on guided image filtering~\cite{song2014real} and feed-forward convolutional neural network~\cite{zhang2015end} are also found to be effective in transferring photos into sketches. A very recent work similar to ours is reported by Zhang \etal \cite{zhang2017content}. They proposed a two-branch FCNN to learn content and texture respectively, and then fused them through a face probability map. Although their results are impressive, their sketch textures do not look natural and the facial components are over smoothed.

%------------------------------------------------------------------------
\subsection{Style Transfer with CNN}
Texture synthesis has long been a challenging task. Traditional methods can only imitate repetitive patterns. Recently, Gatys \etal ~\cite{gatys2015texture,gatys2015neural} studied the use of CNN in style representation, and proposed a method for transferring the style of one image (referred to as the style image) to another (referred to as the content image). In their method, a target style is first computed based on features extracted from the style image using the VGG-Network. An output image is then generated by iteratively updating the content image and minimizing the difference between its style and the target style. Justin \etal \cite{feifei2016} further accelerated this process by learning a feed forward CNN in the training stage. These methods represent styles by a multi-scale Gram matrix of the feature maps. Since the Gram matrix only cares about global statistics, local structures may be destroyed when the style image is very different from the content image. Although this may not be a problem in transferring artistic styles to images, this will definitely produce noticeable artifacts in the face sketch as people are very sensitive to the distortions of the facial features. In \cite{Chen2016Patch}, Chen and Schmidt proposed a different patch based style transfer method which is better at capturing local structures. However, it is still far from satisfactory to be employed in face sketch synthesis. Our style transfer approach is inspired by but different from the above work~\cite{gatys2015texture,gatys2015neural,feifei2016} in that our target style is computed from image patches of many different images rather than from just one single image. 
\begin{figure*}[t]
\centering
\includegraphics[width=0.85\linewidth]{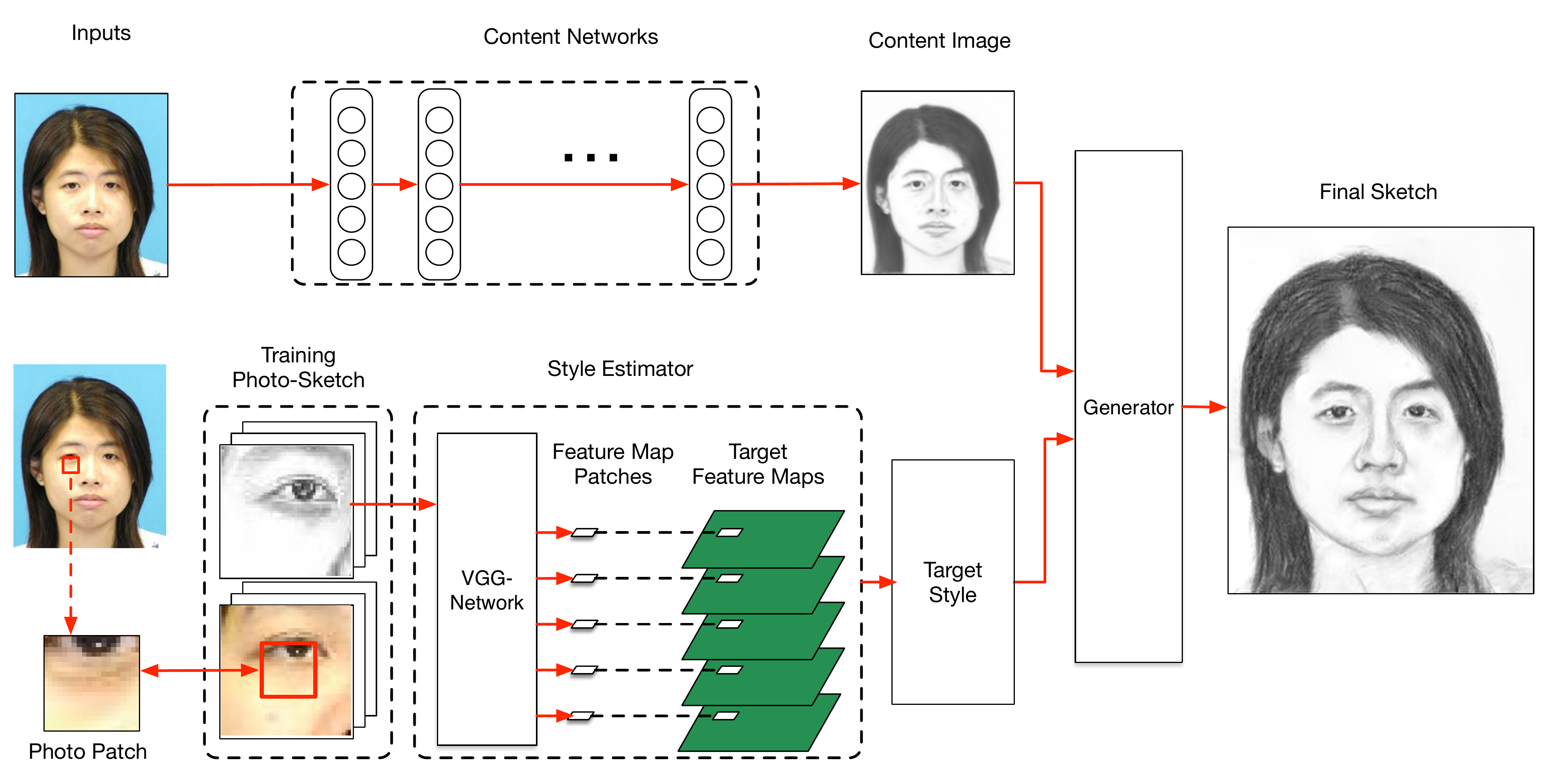}
\caption{The proposed method contains two branches which take an eye-aligned test photo as input. The content network outputs a content image which outlines the face, and the style estimator generates a target style. The final sketch is generated by combing the target style with the content image.}
\label{fig:overview}
\end{figure*}
%==========================================================================

\section{Style Representation}\label{sec:motivation}
Following the work of \cite{gatys2015neural}, we use Gram matrices of VGG-19~\cite{simonyan2014very} feature maps as our style representation. Denote the vectorized $c$th channel of the feature map in the $l$th layer of the final sketch $\mathcal{X}$ by $F^{l}_{c}(\mathcal{X})$. A Gram matrix of the feature map in the $l$th layer is then defined by the inner products between two channels of this feature map, i.e.,
\begin{equation}
G^l_{ij}(\mathcal{X}) = F^l_{i}(\mathcal{X}) \cdot F^l_{j}(\mathcal{X}), 
\label{eq:Gram_element}
\end{equation}
where $G^l(\mathcal{X}) \in {\mathcal{R}^{N_l \times N_l}}$ 
and $N_l$ is the number of channels of the feature map in the $l$th layer. Since $G^l_{ij}(\mathcal{X})$ is an inner product between two channels of the feature map, a Gram matrix is actually a summary statistics of the feature map without any spatial information. Empirically, a Gram matrix of the feature map captures the density distribution of a sketch. For example, if a given style (sketch) image has much less hair than the test photo, the synthesized sketch $\mathcal{X}$ will become brighter than a natural sketch (see experimental results in Section~\ref{subsec:style_transfer}). Thus it is important to have a style (sketch) image which is (statistically) similar to the test photo. Note that, in face sketch synthesis, however, there usually does not exist a single photo-sketch pair in the training set that matches all properties of the test photo. How to compute a target style for the synthesized sketch $\mathcal{X}$ is therefore not trivial, and is the key to the success of this approach. We will introduce a feature-space patch-based approach to solve this problem in Section~\ref{subsec:pyramid_feature_column}.

%==========================================================================
\section{Methodology}
Our method can be classified as a shading synthesis method. The steps of our method are summarized in Fig.~\ref{fig:overview}. First, a preprocessing step as described in~\cite{wang2009face} is carried out to align all photos and sketches in the training set by the centers of the two eyes. An eye-aligned test photo $\mathcal{I}$ is then fed into two branches, namely the content network and the style estimator. The content network converts $\mathcal{I}$ into a content image $\mathcal{C}$, which outlines the shape of the face and the key facial features such as nose, eyes, mouth and hair. The style estimator divides $\mathcal{I}$ into a grid of non-overlapping $16\times16$ patches. For each test patch, it locates the most similar photo patch from the photo-sketch pairs in the training set and produce a target sketch patch from the corresponding sketch in the pair. A pyramid column feature (Section~\ref{subsec:pyramid_feature_column}) is then computed for the target sketch patch. Finally, a target style can be computed from a grid of these pyramid column features, and a final sketch $\mathcal{X}$ can be synthesized by applying the target style to $\mathcal{C}$ through neural style transfer~\cite{gatys2015neural}.  

%------------------------------------------------------------------------
\subsection{Content Image Generation} \label{sec:content_net}
\begin{figure*}[htbp]
\centering
\subfigure[The architecture of content network]{
\label{fig:content_NN_a}\includegraphics[width=0.65\linewidth]{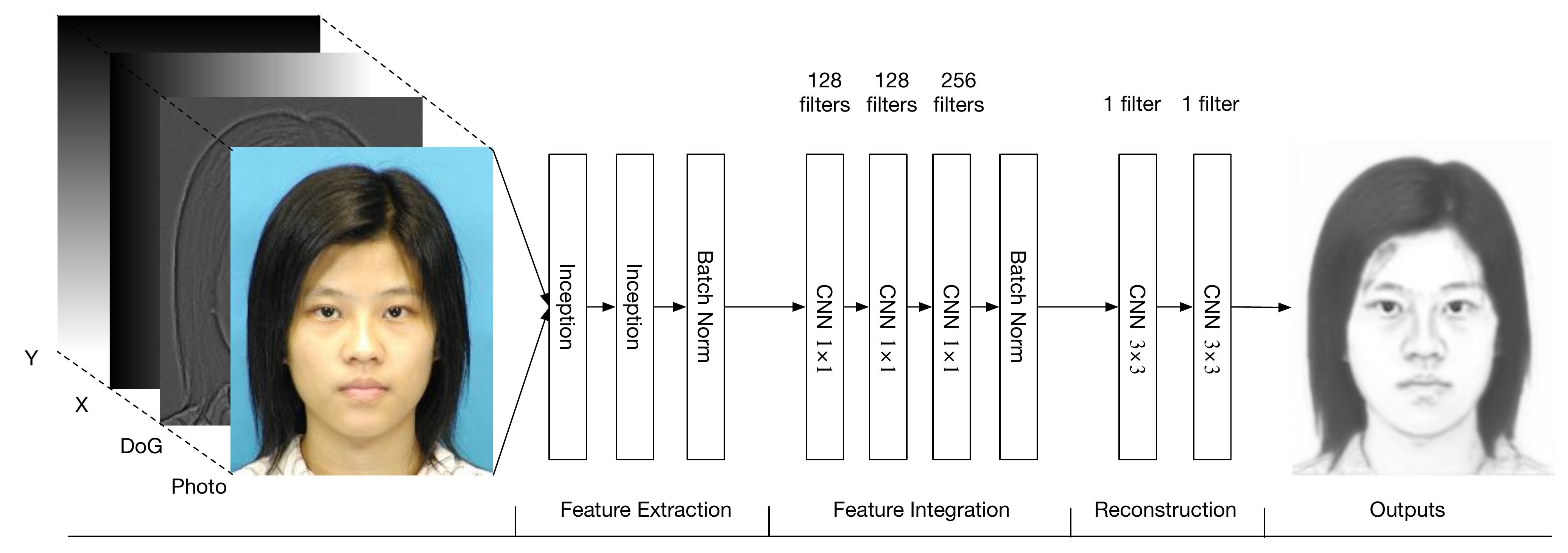}}
\subfigure[Inception module]{
\label{fig:content_NN_b}\includegraphics[width=0.25\linewidth]{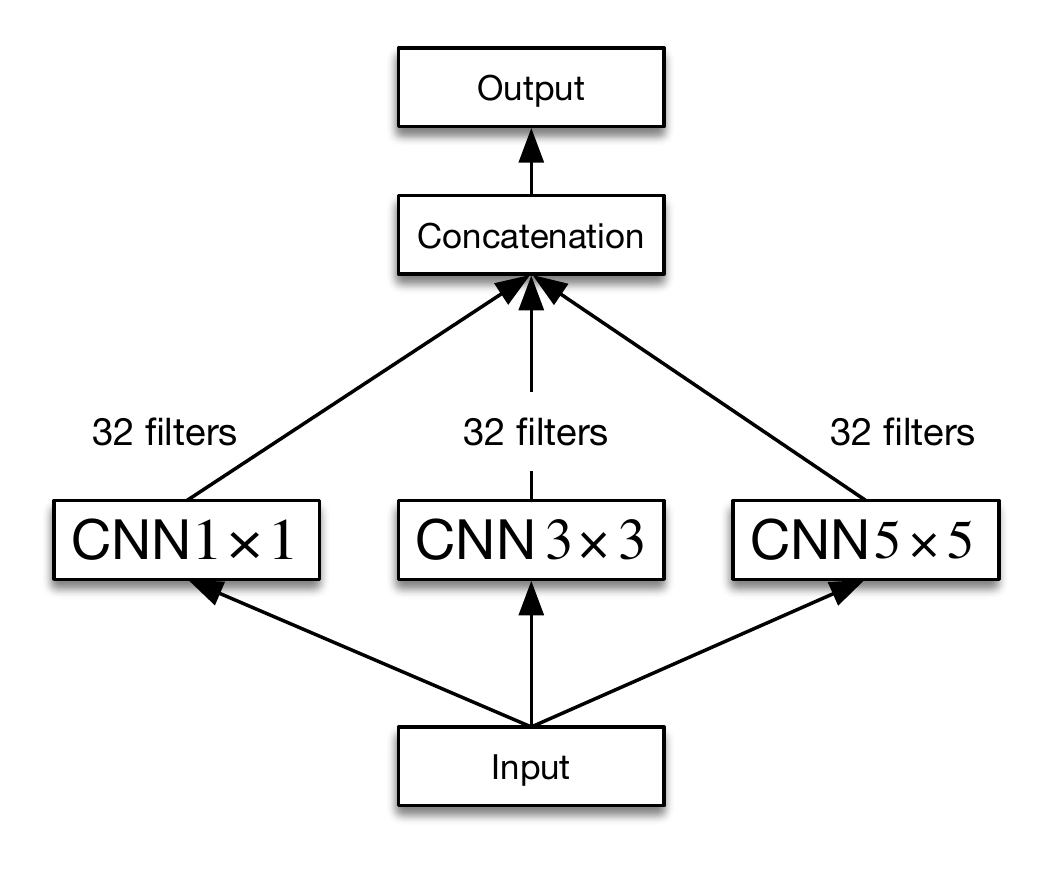}}
\caption{Illustration of the content network for generating a content image. The numbers above the building block denote the number of CNN filters. (a) The architecture of content network. (b) The inception module in (a) contains three groups of filters with different sizes.}
\label{fig:content_NN}
\end{figure*}

The architecture of our content network is shown in Fig.~\ref{fig:content_NN}. Besides the test photo, we feed three extra channels containing the spatial information (i.e., $x$ and $y$ coordinates) and a difference of Gaussian (DoG) image into our content network. As pointed out in \cite{wang2009face}, face sketch synthesis algorithms can benefit from integrating features from multiple resolutions. Hence, we employ an inception module %inspired by the GoogLeNet~
\cite{szegedy2015going} for feature extraction, which concatenates features generated from three groups of filters %with different spatial resolutions. Our inception unit contains 3 different size of filters
with a size of $1\times1$, $3\times3$ and $5\times5$ respectively (see Fig.~\ref{fig:content_NN_b}). Features extracted using a two-layer-inception module are then fed into a three-layer-CNN for feature integration, where all filters have a size of $1\times1$. Finally, the integrated features are used to reconstruct the content image $\mathcal{C}$  by a two-layer-CNN with the filter size being $3\times3$. Since $L_1$-norm is better at preserving details than $L_2$-norm, we use the $L_1$-norm between $\mathcal{C}$ and the ground truth sketch $S$ as the loss function in training our content network, i.e.,
\begin{equation}
\mathcal{L} = \frac{1}{N} \sum \limits_{i=1}^N \|\mathcal{C}_i - S_i\| \label{eq:content-net-loss}
\end{equation}
where $N$ is the number of training photos. 

\subsection{Style Estimation} \label{subsec:pyramid_feature_column}
\begin{figure}[htbp]
\centering
\includegraphics[width=0.85\linewidth]{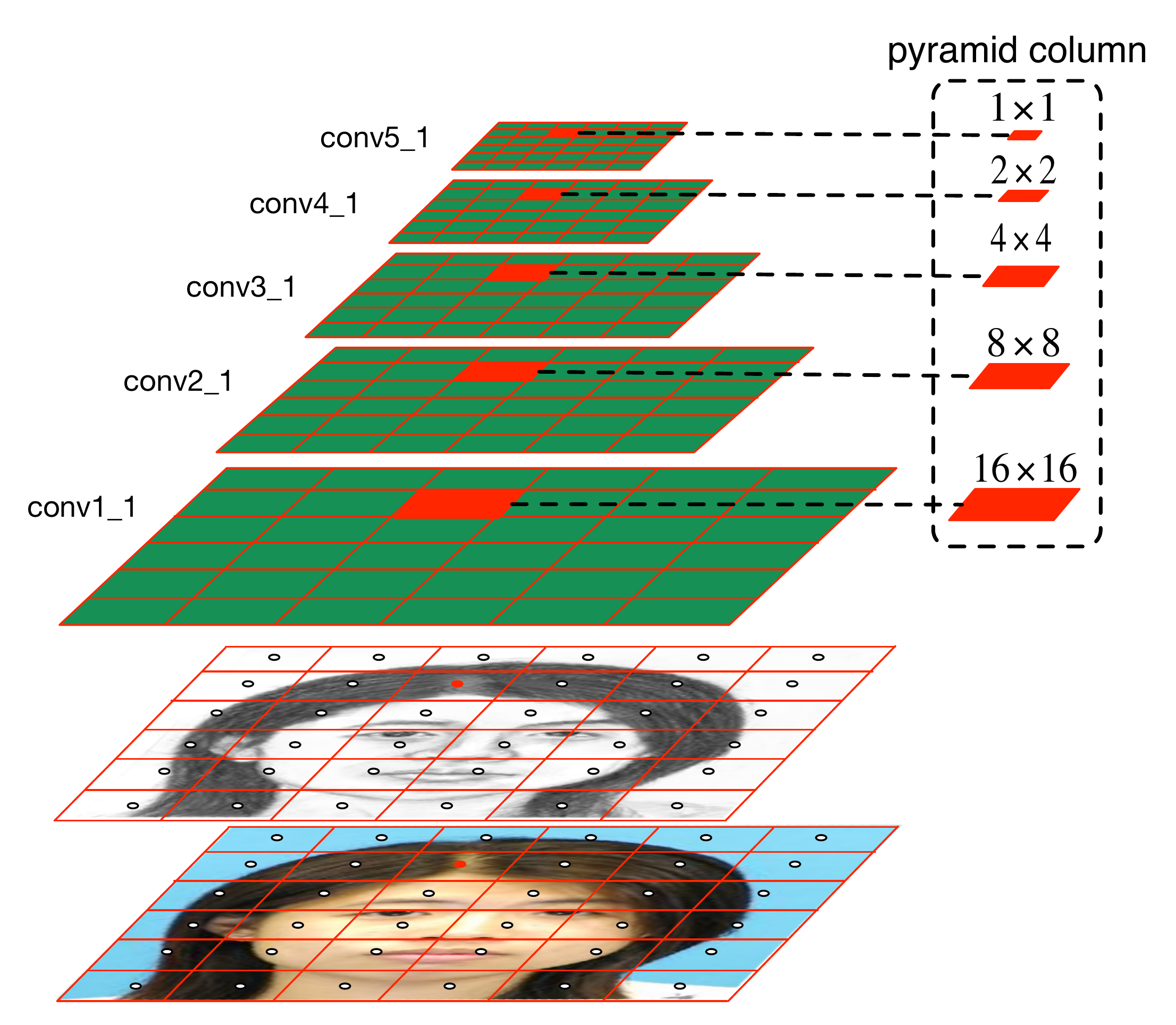}
\caption{Illustration of the \emph{pyramid column feature}. After finding a target sketch patch, we can extract its corresponding feature patches from the original feature maps. These five feature patches make up the \emph{pyramid column feature.}}
\label{fig:pyramidcolumn}
\end{figure}

%------------------------------------------------------------------------
As mentioned previously, there usually does not exist a single photo-sketch pair in the training set that matches all properties of the test photo. In order to estimate a target style for the final sketch $\mathcal{X}$, we subdivide the test photo into a grid of non-overlapping $16\times16$ patches. For each test patch, similar to previous work  \cite{wang2009face,zhou2012markov}, we find the best matching photo patch from the photo-sketch pairs in the training set in terms of mean square error (MSE). A target sketch patch can then be obtained from the corresponding sketch in the photo-sketch pair containing the best matching photo patch. Instead of compositing a style image using the thus obtained target sketch patches, which may show inconsistency across neighboring patches, we adopt a feature-space approach here. We extract feature patches from the feature maps of the original sketch at 5 different layers of the VGG-Network, namely $conv1\_1$, $conv2\_1$, $conv3\_1$, $conv4\_1$ and $conv5\_1$ respectively, that correspond to the target sketch patch. These feature patches have a size of $16\times16$, $8\times8$, $4\times4$, $2\times2$ and $1\times1$ respectively (see Fig.~\ref{fig:pyramidcolumn}). We group these five feature patches of a target sketch patch and call it a  {\em pyramid column feature}. Finally, a target style, in the form of Gram matrices, can be computed directly from a grid of such pyramid column features.

\subsection{Loss Function for Sketch Generation}

Similar to \cite{gatys2015neural}, our loss function is composed of a content loss and a style loss. In addition, we introduce a component loss to enhance the key facial components. The total loss is

\begin{equation}
\mathcal{L}_{t}( \mathcal{X} ) = \alpha \mathcal{L}_{c} + \beta_1 \mathcal{L}_{s} + \beta_2 \mathcal{L}_{k},
\label{eq:Total_loss}
\end{equation}
where $\alpha$, $\beta_1$ and $\beta_2$ are the weights for the different loss terms.
We minimize the loss function by updating the target sketch $\mathcal{X}$ in the same way as \cite{gatys2015neural}. 

The content loss is defined by the difference between the feature map at layer $conv1\_1$ of the synthesized sketch and that of the content image :
\begin{equation}
\mathcal{L}_{c}( \mathcal{X} ) = \| {{F^{\rm{conv1\_1}}}( \mathcal{X} ) - {F^{\rm{conv1\_1}}}( \mathcal{C} )} \|_2^2.
\label{eq:content_loss}
\end{equation}
The style loss is defined by the difference between the Gram matrices of the synthesized sketch and that of the target style:
\begin{equation}
\mathcal{L}_{s} ( \mathcal{X} ) = \sum\limits_{l \in {L_s}} {\frac{1}{{4M_l^2N_l^2}}\| {{G^l}(\mathcal{X} ) - G^l(\mathcal{T})} \|_2^2} 
\label{eq:Gram_loss}
\end{equation}
where $N_l$ denotes the feature map channels at layer $l$, and $M_l$ is the product of width and height of the feature map at layer $l$, and $G^l(\mathcal{T})$ is the Gram matrix computed from the grid of pyramid column features.  

To better transfer styles of the key facial components, we employ a component loss to encourage the key component style of the final sketch to be the same as the target key component style. Since all photos and sketches have been aligned by the centers of the two eyes, the key components lie roughly within a rectangular region $\mathcal{R}$ with the eyes positioned at its upper corners. Here, we define the key component style by Gram matrices computed from feature maps corresponding to  the rectangular region $\mathcal{R}$. The component loss is defined as
\begin{equation}
\mathcal{L}_{k} ( \mathcal{X} ) = \sum\limits_{l \in {L_s}} {\frac{1}{{4\hat{M}_l^2{N}_l^2}}\| {{{\hat G}^l}( \mathcal{X} ) - {\hat G}^l(\mathcal{T})} \|_2^2} 
\label{eq:component_loss}
\end{equation}
where $\hat G^l$ denotes the Gram matrix computed for the rectangular region $\mathcal R$, and $\hat{M}_l$ is the product of width and height of the feature map at layer $l$ corresponding to $\mathcal R$. 

%------------------------------------------------------------------------
\subsection{Implementation Details}

\paragraph*{VGG-19 Parameters} Since the VGG-Network is originally designed for color images, while sketches are gray scale images, we modify the first layer of VGG-Network for gray scale images by setting the filter weights to
\begin{equation}
W^{k} = W^{k}_r+W^{k}_g+W^{k}_b
\label{eq:VGG_weights}
\end{equation}
where $W^{k}_r$, $W^{k}_g$, and $W^{k}_b$ are weights of the $k$th filter in the first convolutional layer for the R, G and B channels respectively, and $W^{k}$ is the weight of the $k$th filter in the first convolutional layer of our modified network.

\paragraph{Data Partition} CUHK \cite{wang2009face} has 88 training photos and 100 test photos, and AR \cite{martinez1998r} has 123 photos. Our training set is composed of the 88 training photos from CUHK and 100 photos from AR. When training the content network, 10\% of the training set are taken out as the validation set. All the 188 photo-sketch pairs are used to generate target sketch.

\paragraph{Training the Content Network} The input photo-sketch pairs are all resized to $288\times288$  and aligned by the centers of the two eyes. A mirror padding is carried out before the convolution operation except when kernel size is $1\times1$ to ensure the output sketch is of the same size as the input. Adadelta \cite{matt2012adadelta} is used as the optimizer because it is stable and much faster than others.  

\paragraph{Sketch Generation} 
In all experiments, we resize the test photos and the photo-sketch pairs in the training set to $288\times288$. The final sketch is obtained by resizing the resulting sketch back to the original size. The size of $\mathcal{R}$ is $48\times48$. The weights in Eq.~(\ref{eq:Total_loss}) are $\alpha=0.004$, $\beta_1=1$ and $\beta_2=0.1$. The minimization is carried out using L-BFGS. Instead of using random noises, we use the content image as a starting point, which will make the optimization process converge much faster. 

\begin{figure*}[htbp]
\centering
\subfigure[Photo]{
\begin{minipage}[b]{0.1\linewidth}
\centering
\includegraphics[width=0.89\linewidth]{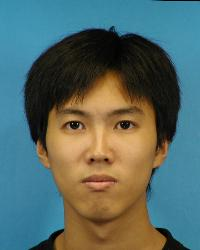}
\includegraphics[width=0.89\linewidth]{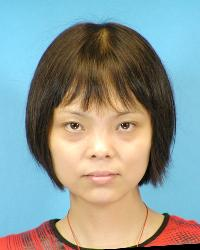}
\includegraphics[width=0.89\linewidth]{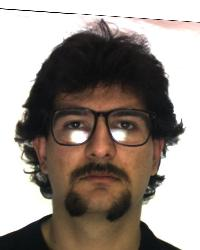}
\includegraphics[width=0.89\linewidth]{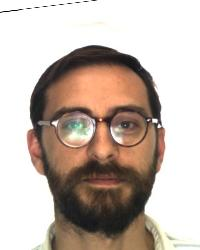}
\end{minipage}
}
\subfigure[MRF]{
\begin{minipage}[b]{0.13\linewidth}
\centering
\includegraphics[width=0.99\linewidth]{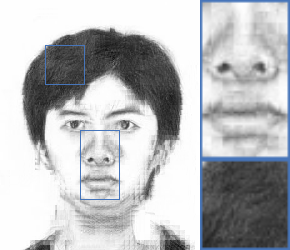}
\includegraphics[width=0.99\linewidth]{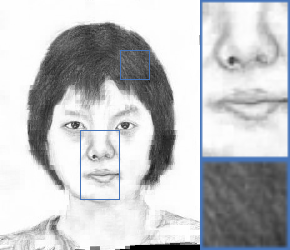}
\includegraphics[width=0.99\linewidth]{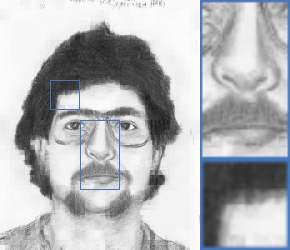}
\includegraphics[width=0.99\linewidth]{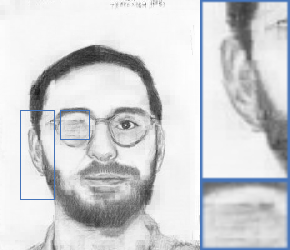}
\end{minipage}
}
\subfigure[WMRF]{
\begin{minipage}[b]{0.13\linewidth}
\centering
\includegraphics[width=0.99\linewidth]{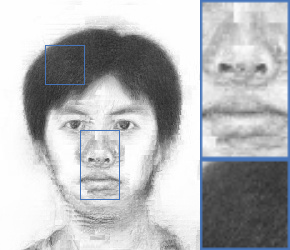}
\includegraphics[width=0.99\linewidth]{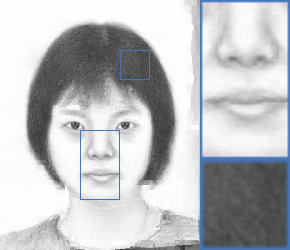}
\includegraphics[width=0.99\linewidth]{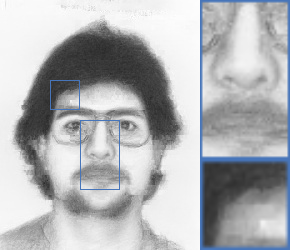}
\includegraphics[width=0.99\linewidth]{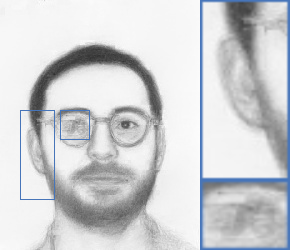}
\end{minipage}
}
\subfigure[SSD]{
\begin{minipage}[b]{0.13\linewidth}
\centering
\includegraphics[width=0.99\linewidth]{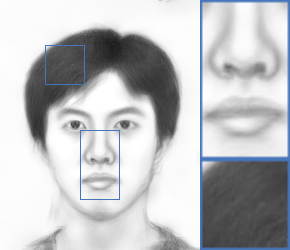}
\includegraphics[width=0.99\linewidth]{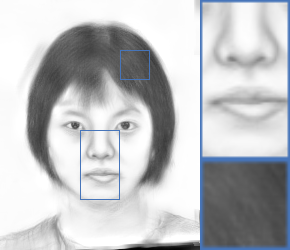}
\includegraphics[width=0.99\linewidth]{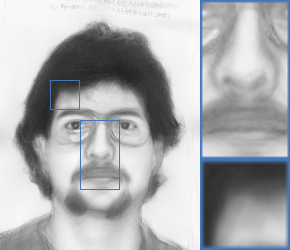}
\includegraphics[width=0.99\linewidth]{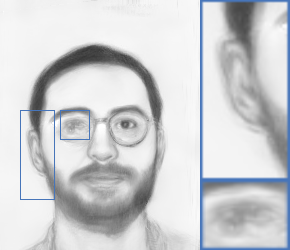}
\end{minipage}
}
\subfigure[FCNN]{
\begin{minipage}[b]{0.13\linewidth}
\centering
\includegraphics[width=0.99\linewidth]{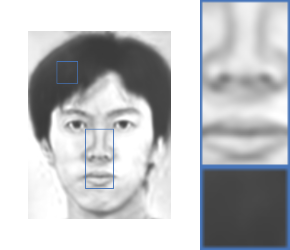}
\includegraphics[width=0.99\linewidth]{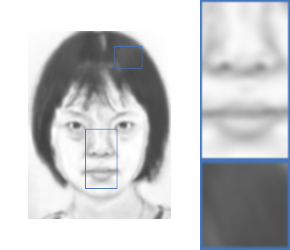}
\includegraphics[width=0.99\linewidth]{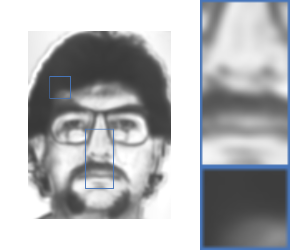}
\includegraphics[width=0.99\linewidth]{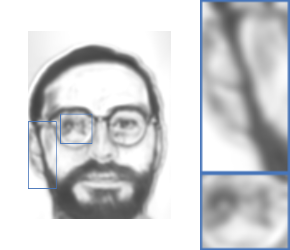}
\end{minipage}
}
\subfigure[BFCN]{
\begin{minipage}[b]{0.13\linewidth}
\centering
\includegraphics[width=0.99\linewidth]{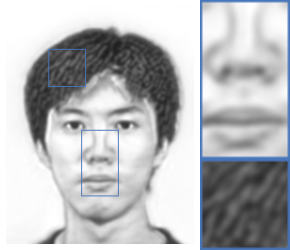}
\includegraphics[width=0.99\linewidth]{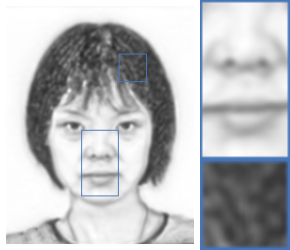}
\includegraphics[width=0.99\linewidth]{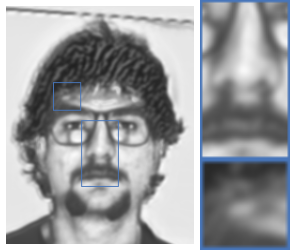}
\includegraphics[width=0.99\linewidth]{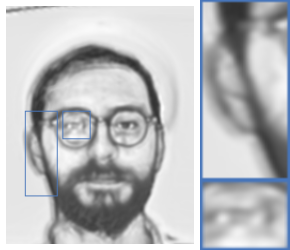}
\end{minipage}
}
\subfigure[Ours]{
\begin{minipage}[b]{0.13\linewidth}
\centering
\includegraphics[width=0.99\linewidth]{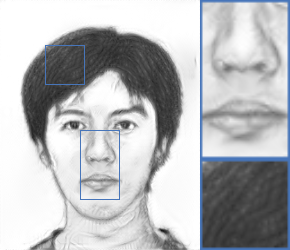}
\includegraphics[width=0.99\linewidth]{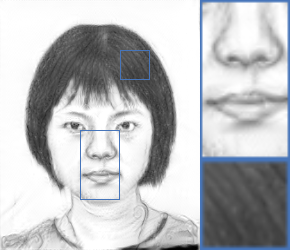}
\includegraphics[width=0.99\linewidth]{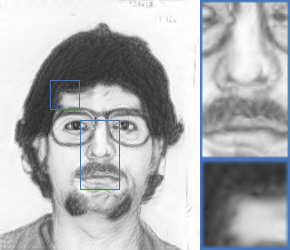}
\includegraphics[width=0.99\linewidth]{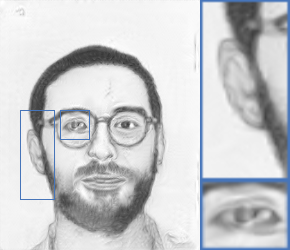}
\end{minipage}
}
\caption{Qualitative evaluation of the face sketch synthesis results on the CUHK (first two rows) and AR (last two rows) datasets. (a) Original Photos. (b) MRF~\cite{wang2009face} (c) WMRF~\cite{zhou2012markov} (d) SSD~\cite{song2014real} (e) FCNN~\cite{zhang2015end} (f) BFCN~\cite{zhang2017content} (g) Ours. The proposed method preserves more texture details (e.g., in the hair and nose), and is also best at keeping the origin structures (e.g., glasses) in the photos.}
\label{fig:qua_eval}
\end{figure*}

%==========================================================================
\section{Experiments}

We evaluate the performance of the proposed method against other state-of-the-art methods on the CUHK student dataset~\cite{wang2009face} and the AR dataset~\cite{martinez1998r}. We compare the results of our method against five other methods, including traditional approaches and recent deep learning models.\del[After discussing the shortcomings of existing quantitative evaluation criteria, we also propose the Normalized Gram Matrix Difference (NGMD) as a new effective metric for quantitatively evaluating the performance of face sketch synthesis.]\blue[After discussing the reason why direct style transfer fails, we evaluate the effectiveness of the proposed \emph{pyramid column feature} and different components of our loss function.]

%------------------------------------------------------------------------
\subsection{Qualitative Evaluation} \label{sec:sketch_gen}

Fig.~\ref{fig:qua_eval} shows the comparison between our methods and five other methods. The first two rows are from the CUHK dataset, and the last two are from the AR dataset. We can see that our method can generate more stylistic sketches than the others. For example, in the hair part, only MRF, BFCN and the proposed method can generate obvious textures. However, the texture in the results of MRF is not continuous across image patches and shows many unpleasant artifacts, whereas the texture in the results of BFCN does not look like natural hand strokes. Both WMRF and SSD introduce an over smoothing effect, and FCNN is not able to produce clear texture. Our method can not only generate textures for hair and moustache, but also shadings (e.g., around the nose).

On the other hand, only FCNN, BFCN and the proposed method can handle structures decorated on the face well, for example the glasses in the last two rows. MRF, WMRF and SSD are exemplar based methods and therefore they cannot handle structures different from the training set. The edges of glasses are not complete in their results. FCNN, BFCN and our method generate the image content by CNN, so they can handle the original structures in the test photo well. However, both FCNN and BFCN cannot generate sharp edges. For example, they produce results which are over smooth in the facial regions around the nose and mouth (see Fig.~\ref{fig:qua_eval}). In contrast, our method can well maintain the image content and create sharp edges.

%------------------------------------------------------------------------
\subsection{Quantitative Evaluation}

\begin{table}[htbp]
\caption{Recognition rate on benchmark datasets. The best performance is colored in red.} \label{tab:reg_percentage}
\small
\begin{tabular}{C{0.2cm}C{0.2cm}C{0.2cm}C{0.1cm}C{0.3cm}C{0.1cm}C{0.35cm}C{0.1cm}C{0.35cm}C{0.1cm}C{0.1cm}C{0.1cm}C{0.25cm}}
\hline
\multicolumn{2}{c|}{\multirow {2}{*}{Methods}} & \multicolumn{5}{c}{AR} & & \multicolumn{5}{c}{CUHK} \\
\cline{3-7} \cline{9-13} \multicolumn{2}{c|}{} & R1 & & R5 & & R10 & & R1 & & R5 & & R10   \\
\hline

\multicolumn{2}{c|}{MRF}  & 97.5\% & & 97.5\% & & \textcolor{red}{100\%} & & 83\% & & 96\% & & 96\% \\
\multicolumn{2}{c|}{WMRF} & 97.5\% & & 97.5\% & & \textcolor{red}{100\%} & & 83\% & & 97\% & & 98\%  \\
\multicolumn{2}{c|}{SSD}  & 96.7\% & & 97.5\% & & \textcolor{red}{100\%} & & \textcolor{red}{87\%} & & 97\% & & 98\%   \\
\multicolumn{2}{c|}{FCNN} & - & & - & & - & & 81\% & & 96\% & & 97\%   \\
\multicolumn{2}{c|}{BFCN} & 92.7\% & & \textcolor{red}{100\%} & & \textcolor{red}{100\%} & & 83\% & & 89\% & & 92\%   \\
\multicolumn{2}{c|}{Ours} & \textcolor{red} {98.4\%} & & 98.4\% & & \textcolor{red} {100\%} & & \textcolor{red} {87\%} & & \textcolor{red} {98\%} & & \textcolor{red} {99\%}   \\
\hline
\end{tabular}
\end{table}

Sketch synthesis methods are commonly evaluated quantitatively  via the face sketch recognition task~\cite{song2014real,wang2009face,zhang2015end,zhou2012markov}. If an algorithm achieves higher sketch recognition rates, it suggests that this method is more effective in synthesizing sketches. We adopt the widely used PCA based recognition method with ``rank-1 (R1)'', ``rank-5 (R5)'' and ``rank-10 (R10)'' criteria~\cite{wang2009face}, where ``rank $n$'' measures the rate of having the correct answer in the top $n$ best matches. The results of different methods are shown in Table~\ref{tab:reg_percentage}. Our method achieves the best performance in all the ``R1'' and ``R5'' and ``R10'' tests. 

%------------------------------------------------------------------------
\subsection{Direct Style Transfer} \label{subsec:style_transfer}

\begin{figure}[htbp]
\centering
\subfigure[]{
\label{fig:style_content} \includegraphics[width=0.23\linewidth]{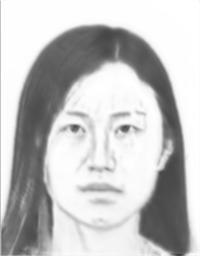}}
\subfigure[]{
\label{fig:style_result1} \includegraphics[width=0.23\linewidth]{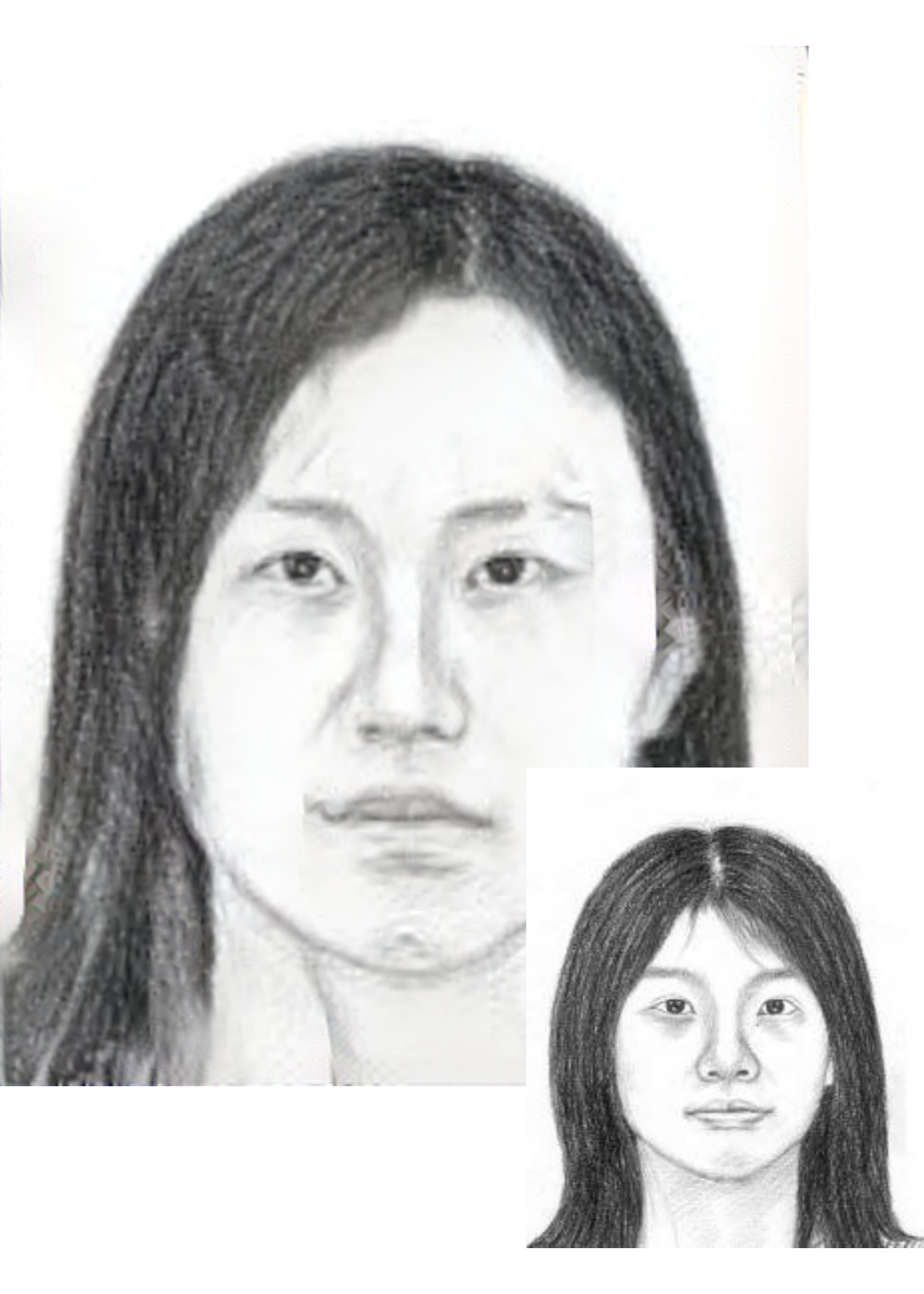}}
\subfigure[]{
\label{fig:style_result2} \includegraphics[width=0.23\linewidth]{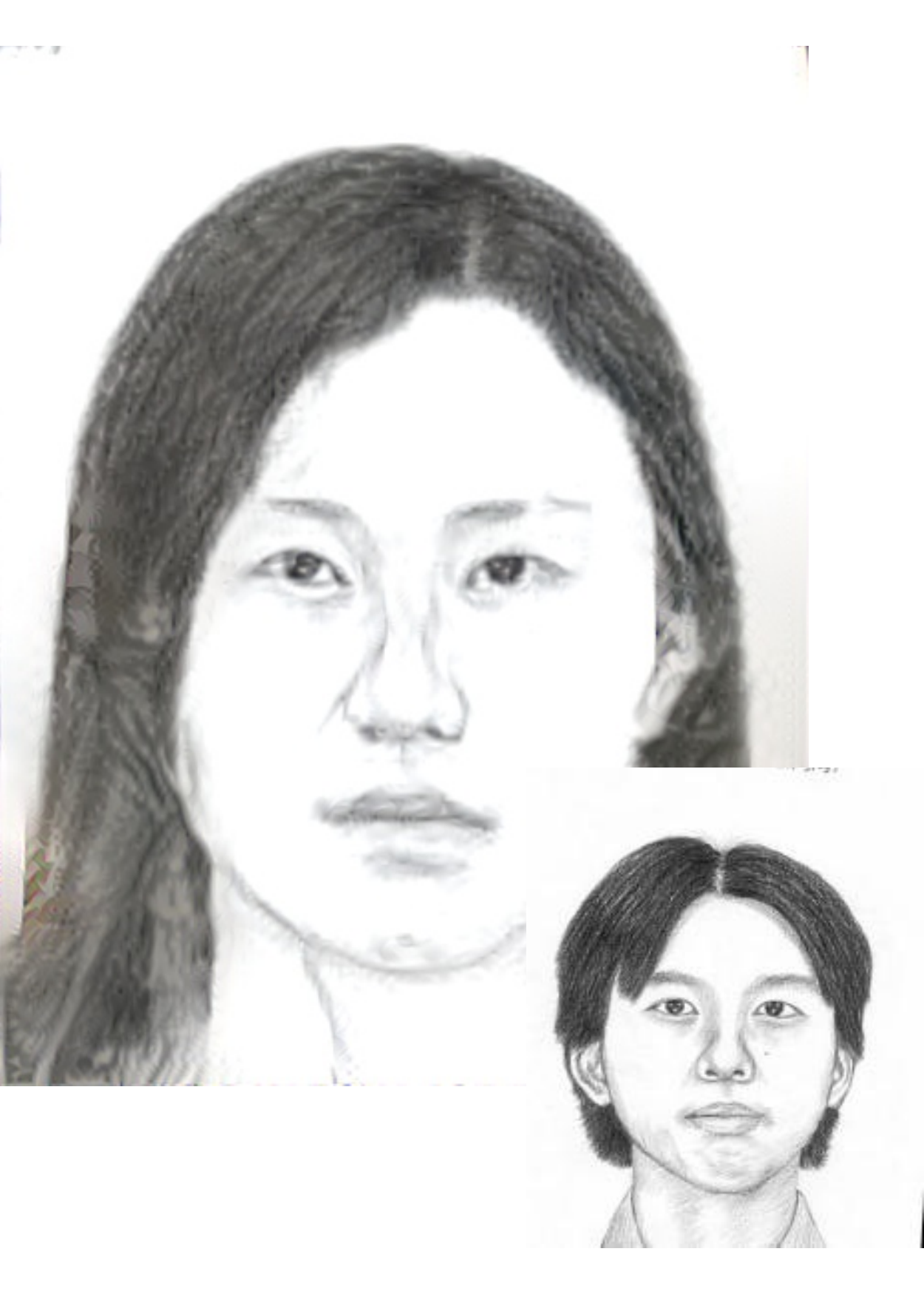}}
\subfigure[]{
\label{fig:style_result3} \includegraphics[width=0.23\linewidth]{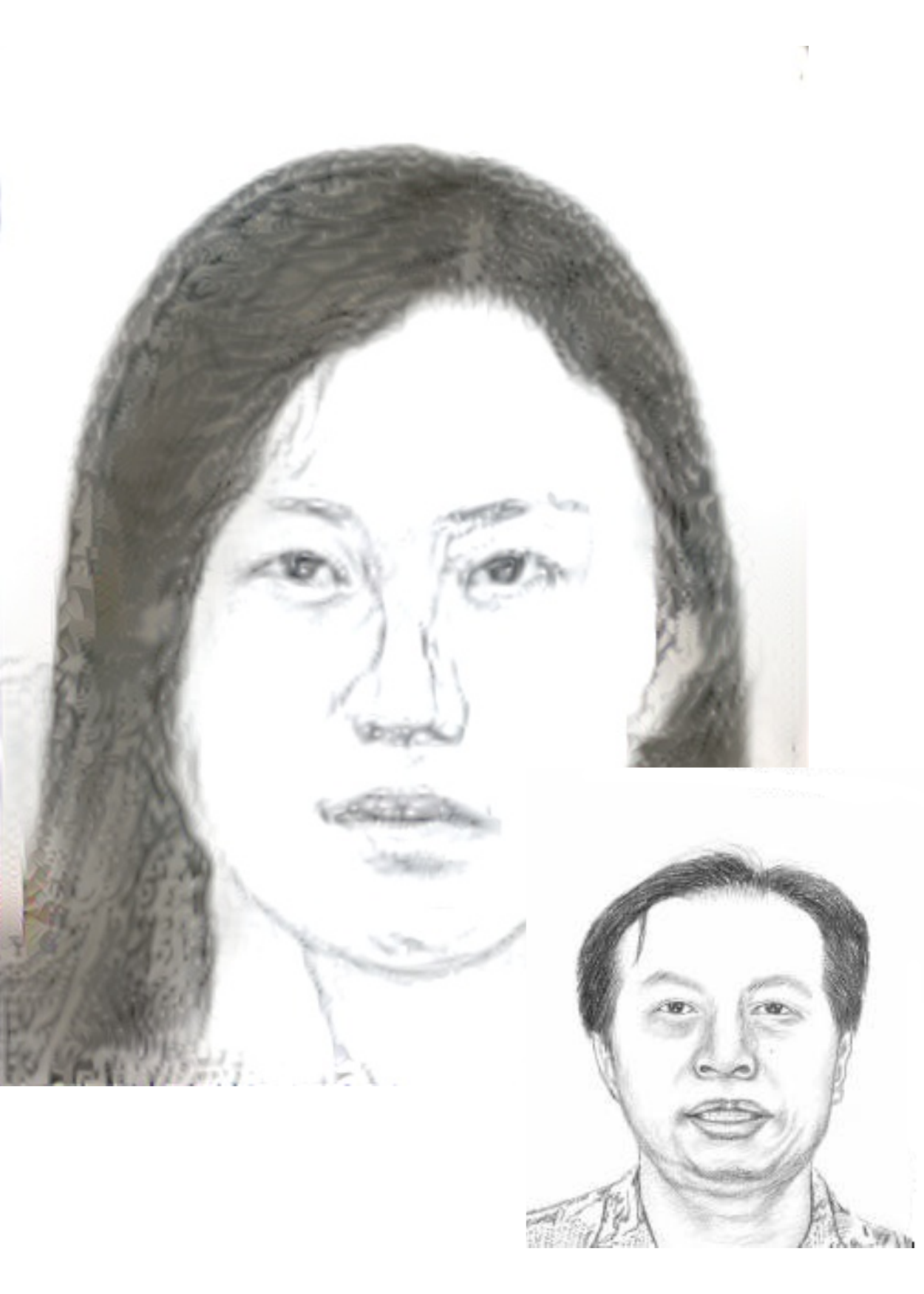}}
\caption{(a) is the content image generated by our content network. (b), (c) and (d) are generated by different styles. It can be seen that the generated sketch becomes brighter as the amount of hair decreases in the style image.}
\label{fig:style_result}
\end{figure}

Although style transfer has shown remarkable performance in transferring artistic style, it cannot be directly applied to face sketch synthesis as the brightness of the generated sketch is easily influenced by the content of the style image (see Fig.~\ref{fig:example_g}). To demonstrate that the Gram matrix captures the density distribution of a sketch, we select 3 sketches with different amount of hair and directly transfer their styles to a content image generated by our content network. The results are shown in Fig.~\ref{fig:style_result}. We can see a clear relationship between the amount of hair in the style image and the pixel intensities of the generated sketch. Some key facial details in Fig.~\ref{fig:style_result2} and \ref{fig:style_result3} are missing due to the overall elevation of pixel intensities. Obviously, direct application of style transfer with an arbitrary style image cannot produce a satisfactory face sketch result. On the other hand, a good looking sketch can be obtained if the style image has a structure similar to the test photo (see Fig.~\ref{fig:style_result1}). This inspires us to introduce our feature-space patch-based method in Section~\ref{subsec:pyramid_feature_column} for generating a compatible target style for our content image. 

\begin{figure*}[htbp]
\centering
\begin{minipage}[t]{0.16\linewidth}
\centering
\includegraphics[width=0.99\linewidth]{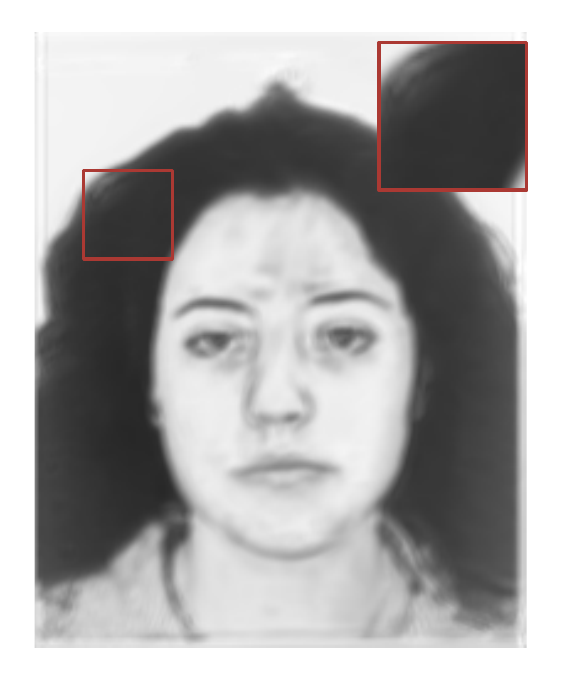}
$\beta_1  = 0 $\\
$\beta_2  = 0 $
\end{minipage}
\begin{minipage}[t]{0.16\linewidth}
\centering
\includegraphics[width=0.99\linewidth]{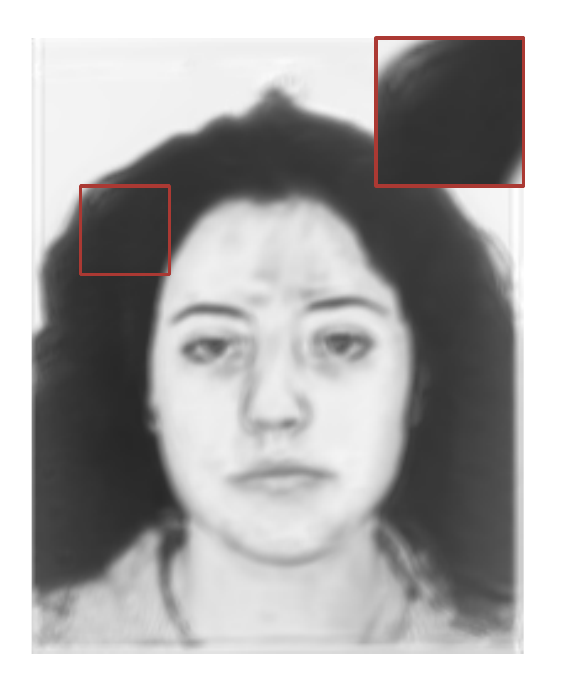}
$\beta_1  = 10^{-3} $\\
$\beta_2  = 10^{-4} $
\end{minipage}
\begin{minipage}[t]{0.16\linewidth}
\centering
\includegraphics[width=0.99\linewidth]{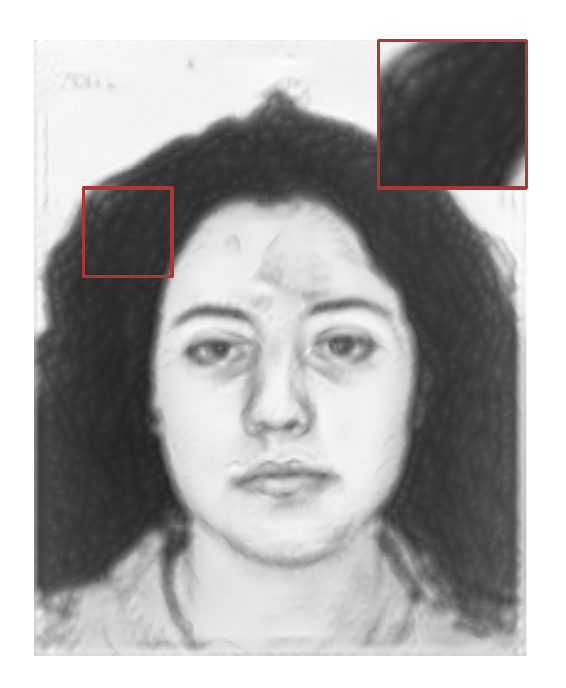}
$\beta_1  = 10^{-2} $\\
$\beta_2  = 10^{-3} $
\end{minipage}
\begin{minipage}[t]{0.16\linewidth}
\centering
\includegraphics[width=0.99\linewidth]{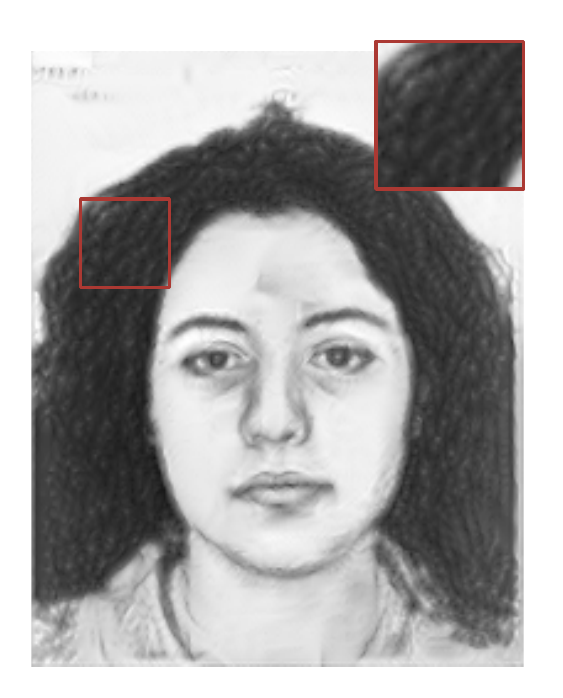}
$\beta_1  = 0.1 $\\
$\beta_2  = 0.01 $
\end{minipage}
\begin{minipage}[t]{0.16\linewidth}
\centering
\includegraphics[width=0.99\linewidth]{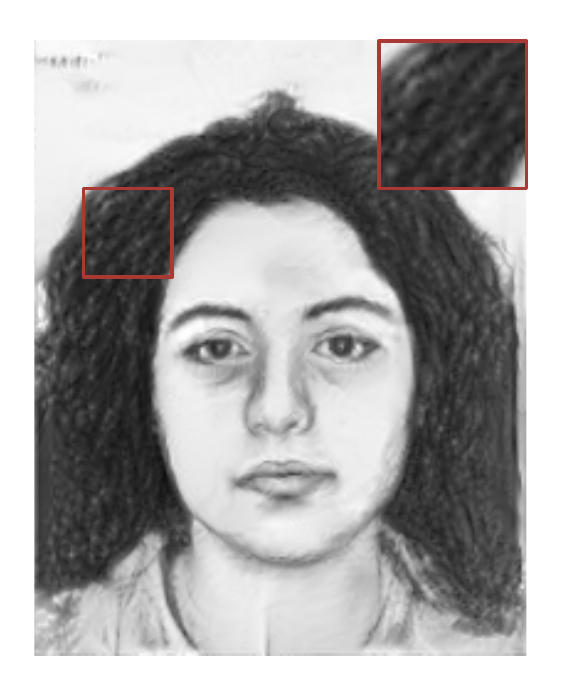}
$\beta_1  = 1 $\\
$\beta_2  = 0.1 $
\end{minipage}
\begin{minipage}[t]{0.16\linewidth}
\centering
\includegraphics[width=0.99\linewidth]{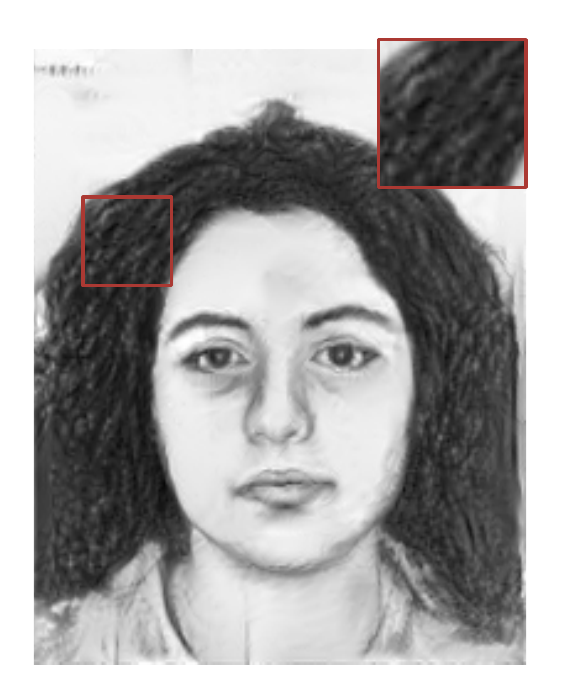}
$\beta_1  = 10^{3} $\\
$\beta_2  = 10^{2} $
\end{minipage}
\caption{Final results for sketch synthesis with fixed $\alpha=0.004$ and different $\beta_1$ and $\beta_2$ values.}
\label{fig:alphg_effect}
\end{figure*}

\begin{figure}[htbp]
\centering
\subfigure[]{
\includegraphics[width=0.22\linewidth]{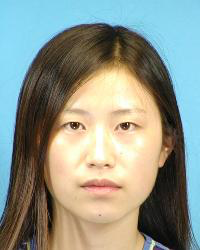}
}
\subfigure[]{
\label{fig:border_a}\includegraphics[width=0.22\linewidth]{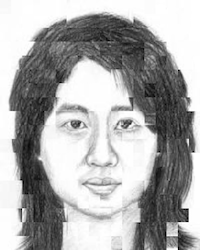}
}
\subfigure[]{
\includegraphics[width=0.22\linewidth]{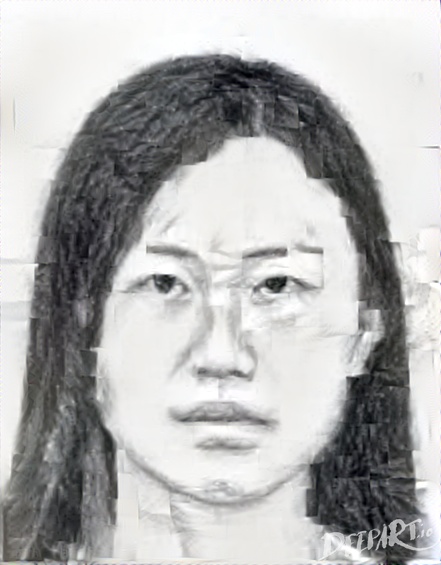}
}
\subfigure[]{
\label{fig:border_c}\includegraphics[width=0.22\linewidth]{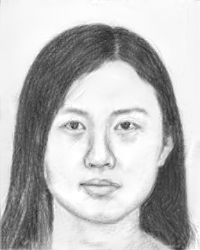}
}
\caption{Feature level style composition helps to eliminate blocky appearance. (a) Photo. (b) A style image composed from target sketch patches. (c) Face sketch obtained from the deep art website using (b) as the style. (d) Face sketch generated using pyramid column feature.}
\label{fig:border_effect}
\end{figure}

%------------------------------------------------------------------------
\blue[\subsection{Effectiveness of the model}]

\paragraph{Pyramid Column Feature} \blue[As demonstrated in the previous section, we need a style image with similar content to the content image in order to apply style transfer for face sketch synthesis. Here, we circumvent this limitation by aggregating pyramid column features of the target sketch patches in estimating the target style. Such a feature space composition is the key to success in our approach. As shown in Fig. \ref{fig:border_effect}, if we simply assemble a target style image from the sketch patches and apply style transfer nav\"iely, the blocky appearance will also be transferred. Although Fig. \ref{fig:border_a} can be improved by making the patches overlap and adding a smoothing term as in \cite{wang2009face,zhou2012markov}, a simpler and more effective way is to compose the target style directly in the feature space, i.e., the proposed \emph{pyramid column feature}. Since the receptive field of the CNN features grow exponentially, the receptive fields of high level feature patches overlap with each other, and this helps to eliminate the blocky appearance problem (see Fig. \ref{fig:border_c}).]

\begin{figure}[htbp]
\centering
\subfigure[$\beta_2=0.001$]{
\includegraphics[width=0.3\linewidth]{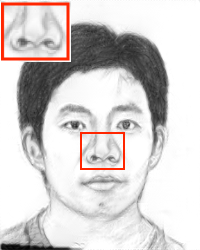}
}
\subfigure[$\beta_2=0.01$]{
\includegraphics[width=0.3\linewidth]{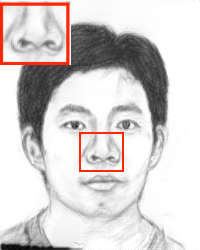}
}
\subfigure[$\beta_2=0.1$]{
\includegraphics[width=0.3\linewidth]{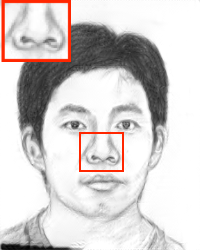}
}
\caption{Comparison between results with different weight of $\mathcal{L}_{k} $ regulation. With the increase of $\beta_2$ the distortion of nose becomes less.}
\label{fig:region_effect}
\end{figure}

\paragraph{Loss Components}
The loss function we minimize during the generation of sketches contains three terms for content, style and key components respectively. The term $\mathcal{L}_{k} $ regularizes the results by encouraging the key component style of the final sketch to be the same as the target key component style. This helps generate better results around the key facial components (see Fig.~\ref{fig:region_effect}). To better understand how style influences the final sketch, we smoothly change the emphasis on style by adjusting $\beta_1$ and $\beta_2$ while keeping $\alpha$  fixed. Fig.~\ref{fig:alphg_effect} shows that the sketch with style transferred contains more textures and looks more like a drawn sketch.

%------------------------------------------------------------------------
\subsection{Generalization}

\begin{figure}[htbp]
\centering
\subfigure[Photo]{
\begin{minipage}[b]{0.22\linewidth}
\centering
\includegraphics[width=0.99\linewidth]{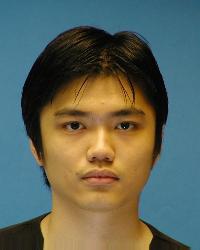}
\includegraphics[width=0.99\linewidth]{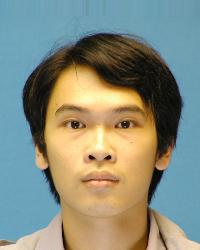}
\includegraphics[width=0.99\linewidth]{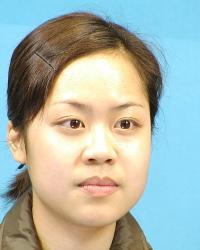}
\includegraphics[width=0.99\linewidth]{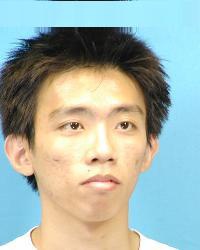}
\end{minipage}
}
\subfigure[ MRF~{\tiny(extended)}]{
\begin{minipage}[b]{0.22\linewidth}
\centering
\includegraphics[width=0.99\linewidth]{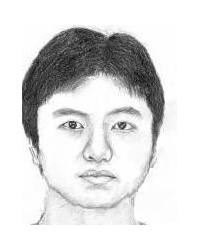}
\includegraphics[width=0.99\linewidth]{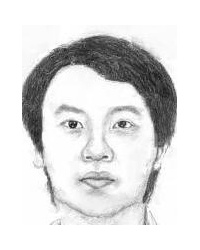}
\includegraphics[width=0.99\linewidth]{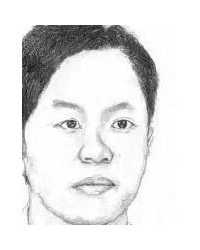}
\includegraphics[width=0.99\linewidth]{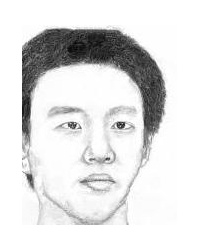}
\end{minipage}
}
\subfigure[BFCN]{
\begin{minipage}[b]{0.22\linewidth}
\centering
\includegraphics[width=0.99\linewidth]{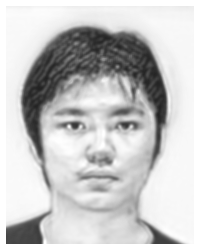}
\includegraphics[width=0.99\linewidth]{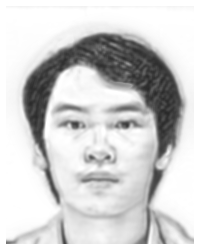}
\includegraphics[width=0.99\linewidth]{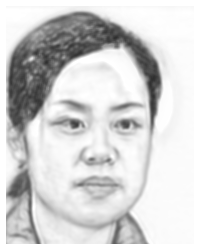}
\includegraphics[width=0.99\linewidth]{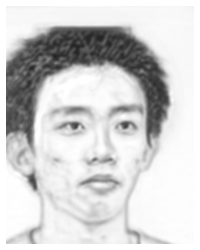}
\end{minipage}
}
\subfigure[Ours]{
\begin{minipage}[b]{0.22\linewidth}
\centering
\includegraphics[width=0.99\linewidth]{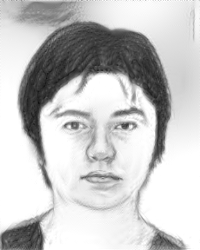}
\includegraphics[width=0.99\linewidth]{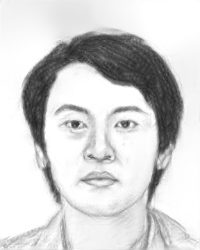}
\includegraphics[width=0.99\linewidth]{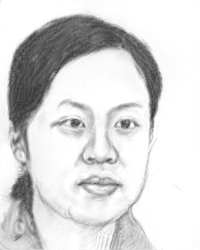}
\includegraphics[width=0.99\linewidth]{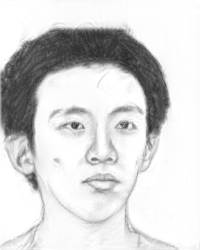}
\end{minipage}
}
\caption{Experiment under different light and pose settings.}
\label{fig:light_pose}
\end{figure}

\begin{figure}[htbp]
\centering
\subfigure[Photo]{
\begin{minipage}[b]{0.22\linewidth}
\centering
\includegraphics[width=0.99\linewidth]{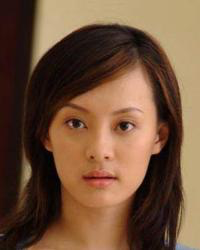}
\includegraphics[width=0.99\linewidth]{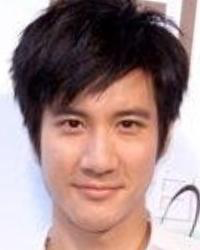}
\includegraphics[width=0.99\linewidth]{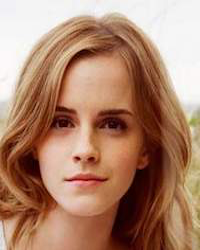}
\includegraphics[width=0.99\linewidth]{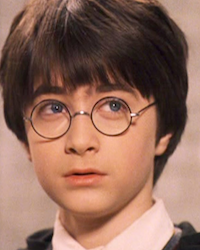}
\end{minipage}
}
\subfigure[FCNN]{
\begin{minipage}[b]{0.22\linewidth}
\centering
\includegraphics[width=0.99\linewidth]{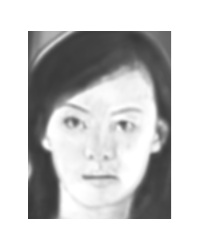}
\includegraphics[width=0.99\linewidth]{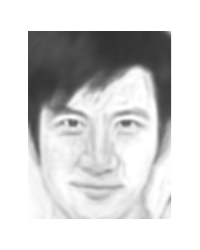}
\includegraphics[width=0.99\linewidth]{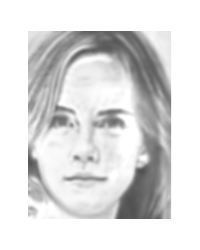}
\includegraphics[width=0.99\linewidth]{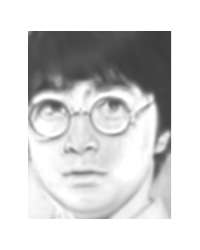}
\end{minipage}
}
\subfigure[BFCN]{
\begin{minipage}[b]{0.22\linewidth}
\centering
\includegraphics[width=0.99\linewidth]{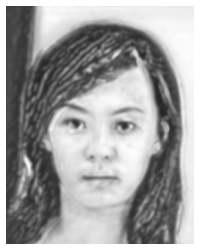}
\includegraphics[width=0.99\linewidth]{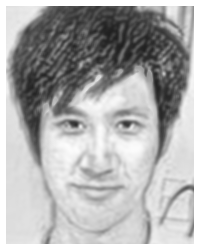}
\includegraphics[width=0.99\linewidth]{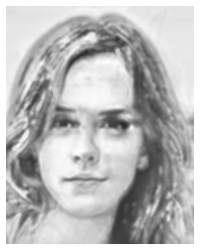}
\includegraphics[width=0.99\linewidth]{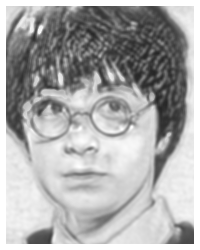}
\end{minipage}
}
\subfigure[Ours]{
\begin{minipage}[b]{0.22\linewidth}
\centering
\includegraphics[width=0.99\linewidth]{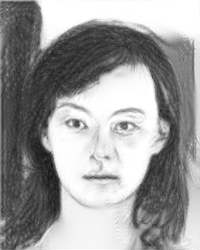}
\includegraphics[width=0.99\linewidth]{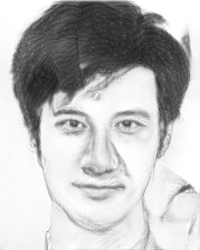}
\includegraphics[width=0.99\linewidth]{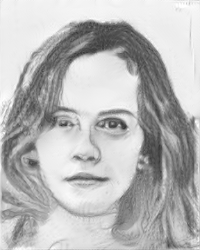}
\includegraphics[width=0.99\linewidth]{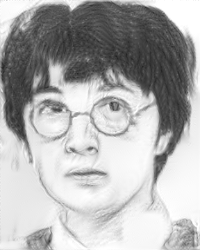}
\end{minipage}
}
\caption{Experiments with real world photos.}
\label{fig:real_world}
\end{figure}

%In this part, we will evaluate the generalization ability of our model in two aspects. 

\paragraph{Light and Pose Invariance} As discussed in \cite{zhang2010lighting}, light and pose changes may influence the result a lot. We choose several photos from \cite{zhang2010lighting} and compare our results with MRF~(extended) \cite{zhang2010lighting} and BFCN \cite{zhang2017content}. As seen from the comparison in Fig.~\ref{fig:light_pose}, our proposed method is not influenced by pose and light changes, and can generate good textures under a laboratory environment. 

\paragraph{Real World Photos} We further test the robustness of our model on some real world photos, and the results are shown in Fig.~\ref{fig:real_world}. The first two rows are Chinese celebrity faces from \cite{zhang2010lighting}, and the latter two come from the world wide web. Since the test photos may not be well aligned, we just turn off the component loss. The parameters we use here are $\alpha=0.004$, $\beta_1=1$, $\beta_2=0$. Although the background is cluttered and the positions of the faces are not strictly constrained, the hair styles of our results are still clear and sharp, whereas FCNN and BFCN fail to produce good textures. 

\section{Discussion}

\blue[The proposed method has two key parts, content image generation and patch matching based sketch feature composition. When the content image is of low quality, the key facial parts of the result may be also unclear, such as the first row in Fig. \ref{fig:real_world}. We believe that training the content network with more data will further improve the result. As for the patch matching part, one concern is that the database may not be suitable for all faces. Nevertheless, our database which contains 188 face sketch pairs can already handle most situations because of the limitation of face content. For example, different shape of glasses are well preserved in Fig. \ref{fig:qua_eval}. Due to the iterative optimization process, the time complexity is another limitation of our method. The Theano implementation of the proposed method takes approximately $60$ seconds to generate a sketch on a GeForce GTX TITAN X platform. The bottle neck lies in the style transfer which requires feeding $\mathcal{X}$ to the VGG-Network to estimate the feature maps and calculate the gradient of Eq.~(\ref{eq:Total_loss}), which is computationally intensive.]

%==========================================================================
\section{Conclusion}

This paper proposes a novel face sketch synthesis method inspired by the process of how artists draw sketches. In our method, the outline of the face is delineated by a content network and the style extracted from sketches drawn by artists are transferred to generate a final sketch. Quantitative evaluations on face sketch recognition demonstrate the effectiveness of the proposed algorithm for face sketch synthesis. Our future work will investigate accelerating technique to reduce the running time and achieve real time face sketch synthesis with style transfer.

\section{Acknowledgment}

We gratefully acknowledge the support of NVIDIA Corporation with the donation of the Titan X Pascal GPU used for this research.

{\small
\bibliographystyle{ieee}
\bibliography{egbib}
}

\end{document}